\documentclass[journal]{IEEEtran}
\usepackage{cite}

\usepackage{booktabs}
\usepackage{graphicx}
\usepackage{xcolor}
\usepackage{amssymb}
\usepackage{amsmath}
\usepackage{tabularx}
\usepackage{adjustbox}
\usepackage{caption}
\usepackage{multirow}
\usepackage{longtable}
\usepackage{hyperref}

\ifCLASSINFOpdf
\else
\fi
\usepackage{array}

\usepackage{stfloats}
\begin{document}
%
\title{A Survey of Graph Neural Networks for Drug Discovery: Recent Developments and Challenges}
%
%
%

\author{\IEEEauthorblockN{Katherine Berry and Liang Cheng}\\
\IEEEauthorblockA{\textit{Department of Electrical Engineering and Computer Science} \\
\textit{University of Toledo}\\
Katherine.Berry@rockets.utoledo.edu, Liang.Cheng@utoledo.edu}
}

\maketitle

\begin{abstract}
Graph Neural Networks (GNNs) have gained traction in the complex domain of drug discovery because of their ability to process graph-structured data such as drug molecule models. This approach has resulted in a myriad of methods and models in published literature across several categories of drug discovery research. This paper covers the research categories comprehensively with recent papers, namely molecular property prediction, including drug-target binding affinity prediction, drug-drug interaction study, microbiome interaction prediction, drug repositioning, retrosynthesis, and new drug design, and provides guidance for future work on GNNs for drug discovery. 
\end{abstract}


%
\IEEEpeerreviewmaketitle

\section{Introduction}

The use of Graph Neural Networks (GNNs) in drug discovery has been explored considerably in recent years, with many novel models evolving to address previous limitations. 
To date, several literature surveys on GNNs for drug discovery have been published. However, they are either a broad discussion of the overall concept or a focused analysis of GNNs for a specific category of drug discovery research. In this work, we survey recent and highly cited papers, a total of 38 research papers as well as the four survey papers, to cover the research categories comprehensively, namely molecular property prediction, including drug-target binding affinity prediction, drug-drug interaction study, microbiome interaction prediction, drug repositioning, retrosynthesis, and new drug design, as shown in Table \ref{tab:GNN_surveys}.

\begin{table*}[!b]
    \centering
    \caption{Overview of Surveys on Graph Neural Networks for Drug Discovery}
    \label{tab:GNN_surveys}
    \renewcommand{\arraystretch}{1.3} 
    \begin{tabular}{p{5cm}>{\centering\arraybackslash}p{1.5cm}>{\centering\arraybackslash}p{1cm}>{\centering\arraybackslash}p{1.5cm}>{\centering\arraybackslash}p{1.5cm}>{\centering\arraybackslash}p{1.7cm}>{\centering\arraybackslash}p{1.5cm}>{\centering\arraybackslash}p{1cm}}
        \toprule
        \textbf{Survey} & \textbf{Molecular Property Prediction} & \textbf{Drug-Target Binding Affinity} & \textbf{Drug-Drug / Synergy} & \textbf{Microbiome Interaction} & \textbf{Drug Repositioning} & \textbf{Retrosynthesis} & \textbf{New Drug Design} \\
        \midrule
        A compact review of molecular property prediction with graph neural networks \cite{Wieder} & \checkmark \\
        \hline
        Graph Neural Networks and Their Current Applications in Bioinformatics \cite{Zhang} & \checkmark & \checkmark & \checkmark \\
        \hline        
        A survey of drug-target interaction and affinity prediction methods via graph neural networks \cite{affsurvey} & & \checkmark \\
        \hline
        A review on graph neural networks for predicting synergistic drug combinations \cite{Besharatifard} & & & \checkmark\\
        \hline
        \textbf{This paper} & \checkmark & \checkmark & \checkmark & \checkmark & \checkmark & \checkmark & \checkmark \\
        \bottomrule
    \end{tabular}
\end{table*}

\subsection{Drug Discovery}
Drug discovery defines the process of identifying new medications for development in the pharmaceutical industry, which often requires consideration of many factors in chemistry and biology. Due to this extensive scope, traditional methods for evaluating candidate drugs are quite costly and time-consuming, and the unfortunate reality of this is that many of these candidate drugs fail at some point in the development process. 


Descriptor-based machine learning methods have been used to interpret information about candidate drug compounds to predict their attributes, and this has been met with promising results. However, the intrinsic graph structure of these compounds creates a unique and nuanced problem domain, with potentially important structural information being left out by these methods. GNNs have been favored in addressing this issue due to their innate ability to represent and extract important features from raw molecule data.

\subsection{GNNs}
In the context of GNNs, a graph is a data structure consisting of vertices (or nodes, e.g. atoms in a molecule) and edges (e.g. a chemical bond in a molecule). Each node has its own feature representation or information such as properties of an atom. The core idea behind GNNs is message passing. GNNs learn by iteratively exchanging and aggregating node and edge information between neighboring nodes. After aggregating messages from its neighbors, each node updates its feature representation by combining the node's previous features with the aggregated information from its neighborhood, often done using another neural network or a gated mechanism. Thus, GNNs learn rich representations (or embeddings) of both node-specific features and the intricate relationships (topology) between nodes within the graph. These embeddings are then passed to a readout function implemented as a neural network layer to perform specific tasks such as node regression, edge classification, link prediction, and graph regression. 

\begin{figure*}[!b]
    \centering
    \includegraphics[width=0.95\linewidth]{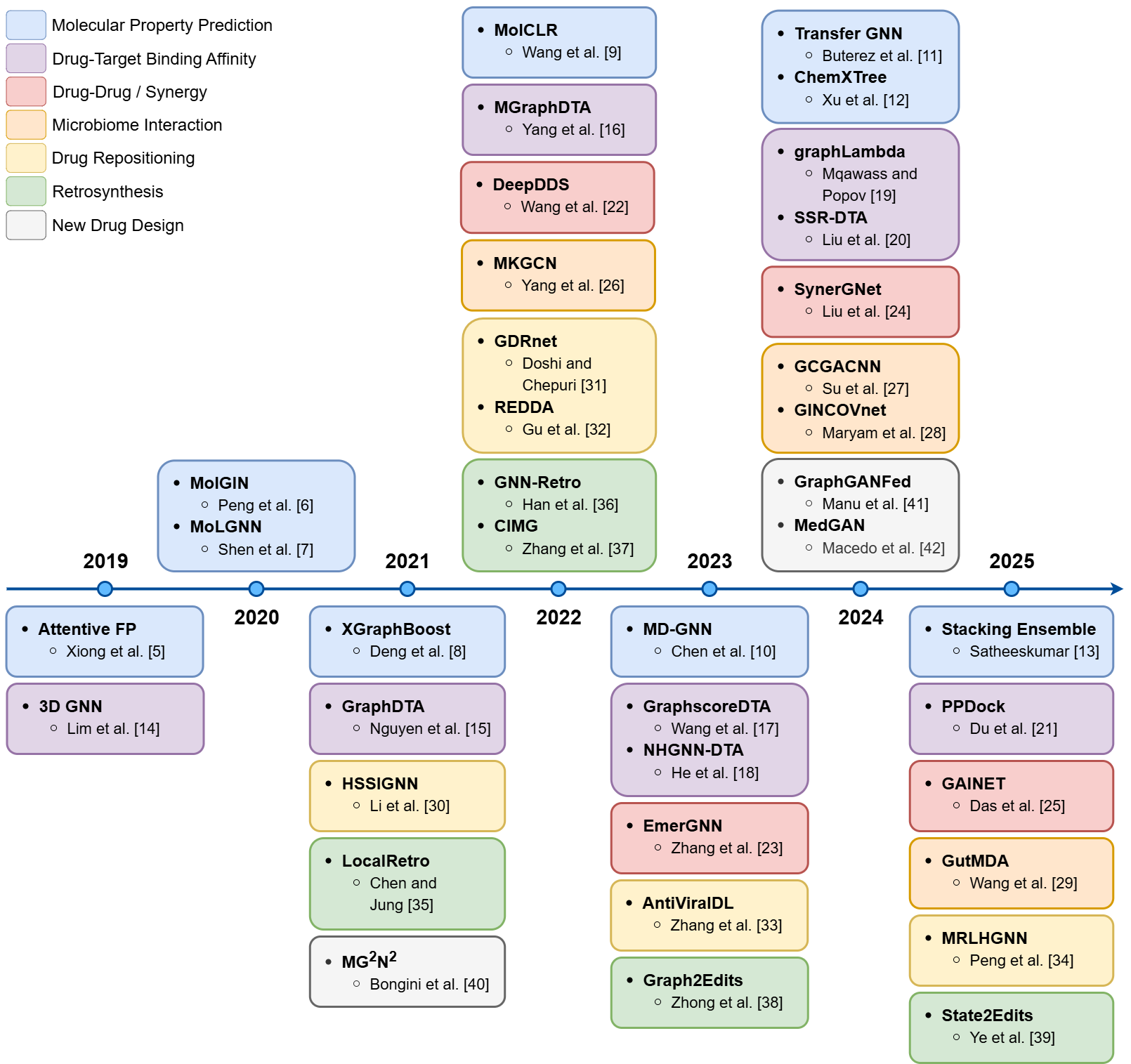}
    \caption{Timeline of Reviewed Work}
    \label{fig:timeline}
\end{figure*}

The reviewed literature has been sectioned into seven categories to create easy-to-follow threads in this paper, namely molecular property prediction, including drug-target binding affinity prediction, drug-drug interaction study, microbiome interaction prediction, drug repositioning, retrosynthesis, and new drug design. A full timeline of the reviewed work is shown in Figure \ref{fig:timeline}, color coded for each category. 

The results and performance of GNNs for drug discovery have progressively improved and shown promising potential with each new contribution, although the state of the art is still evolving to this day. This survey of recent development and challenges will help tackle questions like (i) What model architectures have improved prediction or generation tasks? (ii) What datasets and benchmarks are currently driving this research field? (iii) What are the major challenges facing GNNs for drug discovery today? and (iv) Are GNN models transparent enough for use in regulatory environments?

The rest of this paper is organized as follows. Section \ref{sec:gnn_arc} describes popular GNN architectures used for drug discovery research and the related metrics and datasets. Section \ref{sec:molecularproperty} to Section \ref{sec:newdrug} summarize surveyed papers in those seven categories, respectively. Lastly, future work and conclusions are discussed in Section \ref{sec:futurework} and Section \ref{sec:conclusion}. 

\clearpage

\section{GNN Architecture, Metrics and Datasets for Drug Discovery}\label{sec:gnn_arc}

A summary of the drug discovery research reviewed, including information about GNN models, evaluation metrics, and datasets, is provided in Table \ref{tab:Summary_Review}. 

\subsection{GNN models}
There have been many GNN architectures or models used in the research surveyed. Some popular ones are listed below.
\begin{itemize}
    \item GCN (Graph Convolutional Network, 2017): update node’s representation by aggregating feature information from its neighbors (1-hop, 2-hops, or multi-hops).
    \item GAT (Graph Attention Network, 2018): assign different attention weights to different neighbors and focusing more on relevant nodes during aggregation.
    \item GIN (Graph Isomorphism Network, 2019): use a sum aggregator to capture neighbor features without loss of information combined with an MLP to increase model capacity for representation learning.
    \item MPNN (Message Passing Neural Network, 2017): iteratively pass messages containing node and connection information between neighboring nodes in a graph, which are used to update node representations. A Directed Message Passing Neural Network (DMPNN) is a type of MPNN for graphs with directed edges.
\end{itemize}

\begin{table*}[htbp]
\caption{Summary of the Literature Review}
\label{tab:Summary_Review}
\centering
\small
\begin{adjustbox}{width=\textwidth}
\begin{tabular}{@{}l l l l l l @{}}
\toprule
\textbf{Category} & \textbf{Reference} & \textbf{Proposed Model} & \textbf{Architecture} & \textbf{Metrics} & \textbf{Dataset(s)} \\
\midrule
Molecular Property Prediction & Xiong et al. (2019) \cite{AttentiveFP} & Attentive FP & GAT & \begin{tabular}[t]{@{}l@{}}RMSE, PRC, ROC \end{tabular} & \begin{tabular}[t]{@{}l@{}}Solubility, malaria bioactivity, photovoltaic efficiency \\ESOL, FreeSolv, Lipop, MUV, HIV, BACE, BBBP \\Tox21, ToxCast, SIDER, ClinTox \end{tabular} \\

& Peng et al. (2020) \cite{MolGIN} & MolGIN & GIN & RMSE, AUC & \begin{tabular}[t]{@{}l@{}} LogD$_{7.4}$, LogS, LD50, PPB, P-gp inhibitors, Ames, \\Tox21 \end{tabular}\\

& Shen et al. (2020) \cite{MoLGNN} & MoLGNN & GINVAE & ROC-AUC & \begin{tabular}[t]{@{}l@{}}BACE, BBBP, SIDER, ClinTox, HIV, JAK 1, JAK 2 \\JAK 3 \end{tabular}\\

& Deng et al. (2021) \cite{XGraphBoost} & XGraphBoost & \begin{tabular}[t]{@{}l@{}}GCN \\GGNN \\DMPNN \end{tabular} & \begin{tabular}[t]{@{}l@{}}RMSE, AUC \end{tabular} & \begin{tabular}[t]{@{}l@{}}BBBP, BACE, ClinTox, HIV, SIDER, Tox21, ToxCast \\Lipop, FreeSolv, ESOL \end{tabular} \\

& Wang et al. (2022)\cite{MolCLR} & MolCLR & \begin{tabular}[t]{@{}l@{}}GCN \\GIN \end{tabular} & \begin{tabular}[t]{@{}l@{}}ROC-AUC, RMSE \end{tabular} & \begin{tabular}[t]{@{}l@{}}BBBP, Tox21, ClinTox, HIV, BACE, SIDER, MUV \\FreeSolv, ESOL, Lipop, QM7, QM8, QM9 \end{tabular} \\

& Chen et al. (2023) \cite{MDGNN} & MD-GNN & \begin{tabular}[t]{@{}l@{}}GCN \\GAT \end{tabular} & MAE & PubChem \\

& Buterez et al. (2024) \cite{Buterez} & Transfer GNN & VGAE & \begin{tabular}[t]{@{}l@{}}R$^2$, MAE, AUROC, MCC \end{tabular} & \begin{tabular}[t]{@{}l@{}}AstraZeneca, MF-PCBA, QM7b, QMugs \end{tabular} \\

& Xu et al. (2024) \cite{ChemXTree} & ChemXTree & MPNN & \begin{tabular}[t]{@{}l@{}} ROC-AUC \end{tabular} & \begin{tabular}[t]{@{}l@{}}BBBP, BACE, HIV, ClinTox, AMES, \\CYP3A4\_Substrate, CYP3A4\_inhibitor, \\CYP2D6\_Substrate, CYP2D6\_inhibitor, \\CYP2C9\_Substrate, CYP2C9\_inhibitor, \\Bioavailability \end{tabular} \\

& Satheeskumar (2025) \cite{StackingEnsemble} & Stacking Ensemble & GNN & \begin{tabular}[t]{@{}l@{}}R$^2$, MAE, RMSE, Pearson, \\Spearman \end{tabular} & \begin{tabular}[t]{@{}l@{}}ChEMBL \end{tabular} \\
\midrule

Drug-Target Binding Affinity & Lim et al. (2019) \cite{Lim} & 3D GNN & GAT & \begin{tabular}[t]{@{}l@{}}AUROC, adjusted LogAUC\\PRAUC, sensitivity \\specificity, balanced accuracy \end{tabular} & \begin{tabular}[t]{@{}l@{}}DUD-E, PDBbind, ChEMBL, MUV \end{tabular} \\

& Nguyen et al. (2021) \cite{GraphDTA} & GraphDTA & \begin{tabular}[t]{@{}l@{}}GCN \\GAT \\GIN \\GAT-GCN \end{tabular} & \begin{tabular}[t]{@{}l@{}}CI, MSE \end{tabular} & \begin{tabular}[t]{@{}l@{}}Davis, KIBA \end{tabular} \\

& Yang et al. (2022) \cite{MGraphDTA} & MGraphDTA & \begin{tabular}[t]{@{}l@{}}MGNN \\MCNN \end{tabular} & \begin{tabular}[t]{@{}l@{}}Precision, Recall, AUC, MSE, \\CI, $r_m^2$ index, Spearman \end{tabular} & \begin{tabular}[t]{@{}l@{}}Davis, Filtered Davis, KIBA, Metz \end{tabular} \\

& Wang et al. (2023) \cite{GraphScoreDTA} & GraphscoreDTA & GNN-GRU & \begin{tabular}[t]{@{}l@{}}R, CI, RMSE, MAE, SD \end{tabular} & \begin{tabular}[t]{@{}l@{}}PDBbind, CASF2016, CASF2013 \end{tabular}\\

& He et al. (2023) \cite{NHGNNDTA} & NHGNN-DTA & \begin{tabular}[t]{@{}l@{}}HGNN \\GIN \end{tabular} & \begin{tabular}[t]{@{}l@{}}MSE, CI, $r_m^2$ \end{tabular} & \begin{tabular}[t]{@{}l@{}}Davis, KIBA \end{tabular} \\

& Mqawass and Popov (2024) \cite{graphLambda} & graphLambda & \begin{tabular}[t]{@{}l@{}}GCN \\GAT \\GIN \end{tabular} & \begin{tabular}[t]{@{}l@{}}R, RMSE \end{tabular} & PDBbind, CASF16, CSAR HiQ NRC \\

& Liu et al. (2024) \cite{SSRDTA} & SSR-DTA & \begin{tabular}[t]{@{}l@{}}GAT \\BiGNN \end{tabular} & \begin{tabular}[t]{@{}l@{}}MSE, CI, Pearson, $r_m^2$ \end{tabular} & \begin{tabular}[t]{@{}l@{}}Davis, KIBA, Metz, BindingDB \end{tabular} \\

& Du et al. (2025) \cite{PPDock} & PPDock & EGNN & \begin{tabular}[t]{@{}l@{}}Ligand RMSD, \\Centroid Distance \end{tabular} & PDBbind \\
\midrule

Drug-Drug / Synergy & Wang et al. (2022) \cite{DeepDDS} & DeepDDS & \begin{tabular}[t]{@{}l@{}}GAT \\GCN \end{tabular} & \begin{tabular}[t]{@{}l@{}}ROC AUC, PR AUC, ACC, \\BACC, PREC, TPR, Kappa \end{tabular} & \begin{tabular}[t]{@{}l@{}}DrugBank, CCLE, AstraZeneca \end{tabular} \\

& Zhang et al. (2023) \cite{EmerGNN} & EmerGNN & GNN & \begin{tabular}[t]{@{}l@{}}F1-Score, Accuracy, Kappa, \\PR-AUC, ROC-AUC \end{tabular} & \begin{tabular}[t]{@{}l@{}}DrugBank, TWOSIDES \end{tabular} \\

& Liu et al. (2024) \cite{SynerGNet} & SynerGNet & \begin{tabular}[t]{@{}l@{}}GCN \\JK-Net \end{tabular} & \begin{tabular}[t]{@{}l@{}}AUC, BAC, PPV, FPR, MCC \end{tabular} & \begin{tabular}[t]{@{}l@{}}StringDB, BioGRID, DIP, HPRD, IntAct, MINT, \\MPPI-MIPS, UniProt, AZ-DREAM Challenges, \\GO database, DrugCombDB \end{tabular} \\

& Das et al. (2025) \cite{GAINET} & GAINET & GAT & \begin{tabular}[t]{@{}l@{}}Accuracy, ROC-AUC, AUPRC, \\F1-Score, Recall, Precision \end{tabular} & \begin{tabular}[t]{@{}l@{}}DrugBank \end{tabular} \\
\midrule

Microbiome Interaction & Yang et al. (2022) \cite{MKGCN} & MKGCN & \begin{tabular}[t]{@{}l@{}}GCN \end{tabular} & \begin{tabular}[t]{@{}l@{}}AUC, AUPR \end{tabular} & \begin{tabular}[t]{@{}l@{}}MDAD, aBiofilm, DrugVirus \end{tabular} \\

& Su et al. (2024) \cite{GCGACNN} & GCGACNN & \begin{tabular}[t]{@{}l@{}}GCN \\GAT \end{tabular} & \begin{tabular}[t]{@{}l@{}}AUC, AUPR, F1-Score, Accuracy \end{tabular} & \begin{tabular}[t]{@{}l@{}}MDAD \end{tabular} \\

& Maryam et al. (2024) \cite{Maryam} & GINCOVnet & \begin{tabular}[t]{@{}l@{}}GCN \\GAT \\GINCOV\end{tabular} & \begin{tabular}[t]{@{}l@{}}Balanced accuracy, \\Weighted precision and recall, \\Sensitivity, Specificity, AUROC, \\MCC, F1 \end{tabular} & Collected from literature \\

& Wang et al. (2025) \cite{GutMDA} & GutMDA & GCN & \begin{tabular}[t]{@{}l@{}}AUC, AUPR, F1-Score, \\Accuracy, Specificity, Recall, \\Precision \end{tabular} & \begin{tabular}[t]{@{}l@{}}MASI, MDAD, aBiofilm, DrugVirus \end{tabular} \\
\midrule

Drug Repositioning & Li and Pan (2021) \cite{HSSIGNN} & HSSIGNN & \begin{tabular}[t]{@{}l@{}}GNN \end{tabular} & \begin{tabular}[t]{@{}l@{}}AUC, AUPR \end{tabular} & \begin{tabular}[t]{@{}l@{}}Fdataset, Cdataset \end{tabular} \\

& Doshi and Chepuri (2022) \cite{GDRnet} & GDRnet & \begin{tabular}[t]{@{}l@{}}SIGN \end{tabular} & \begin{tabular}[t]{@{}l@{}}AUPRC, AUROC \end{tabular} & \begin{tabular}[t]{@{}l@{}}DrugBank, Hetionet, GNBR, STRING, IntAct, DGId \end{tabular} \\

& Gu et al. (2022) \cite{REDDA} & REDDA & \begin{tabular}[t]{@{}l@{}}GCN \end{tabular} & \begin{tabular}[t]{@{}l@{}}AUC, AUPR, F1-Score, \\accuracy, recall, precision \end{tabular} & \begin{tabular}[t]{@{}l@{}}Fdataset, Cdataset, KEGG, DrugBank, CTD, STRING, \\UniProt, B-dataset \end{tabular} \\

& Zhang et al. (2023) \cite{AntiViralDL} & AntiViralDL & \begin{tabular}[t]{@{}l@{}}LightGCN \end{tabular} & \begin{tabular}[t]{@{}l@{}}AUC, AUPR \end{tabular} & \begin{tabular}[t]{@{}l@{}}DrugVirus2, US FDA \end{tabular} \\

& Peng et al. (2025) \cite{MRLHGNN} & MRLHGNN & \begin{tabular}[t]{@{}l@{}}GNN \end{tabular} & \begin{tabular}[t]{@{}l@{}}AUC, AUPR, F1-Score, \\Accuracy, Recall, Specificity, \\Precision \end{tabular} & \begin{tabular}[t]{@{}l@{}}Fdataset, Cdataset, KEGG, CTD, DrugBank, STRING, \\UniProt \end{tabular} \\
\midrule

Retrosynthesis & Chen and Jung (2021) \cite{LocalRetro} & LocalRetro & \begin{tabular}[t]{@{}l@{}}MPNN \\GRA \end{tabular} & \begin{tabular}[t]{@{}l@{}}Top-k exact match accuracy, \\Top-k round-trip accuracy \end{tabular} & \begin{tabular}[t]{@{}l@{}}USPTO-50K, USPTO-MIT \end{tabular} \\

& Han et al. (2022) \cite{GNNRetro} & GNN-Retro & \begin{tabular}[t]{@{}l@{}}GNN \end{tabular} & \begin{tabular}[t]{@{}l@{}}Success Rate \end{tabular} & \begin{tabular}[t]{@{}l@{}}USPTO \end{tabular} \\

& Zhang et al. (2022) \cite{CIMG} & CIMG & \begin{tabular}[t]{@{}l@{}}MPNN \end{tabular} & \begin{tabular}[t]{@{}l@{}}Top-k accuracy, ROC \end{tabular} & \begin{tabular}[t]{@{}l@{}}Collected from literature \end{tabular} \\

& Zhong et al. (2023) \cite{Graph2Edits} & Graph2Edits & \begin{tabular}[t]{@{}l@{}}DMPNN \end{tabular} & \begin{tabular}[t]{@{}l@{}}Top-k exact match accuracy, \\Top-k round-trip accuracy \\MaxFrag accuracy \end{tabular} & \begin{tabular}[t]{@{}l@{}}USPTO-50K \end{tabular} \\

& Ye et al. (2025) \cite{State2Edits} & State2Edits & \begin{tabular}[t]{@{}l@{}}DMPNN \end{tabular} & \begin{tabular}[t]{@{}l@{}}Top-k exact match accuracy \end{tabular} & \begin{tabular}[t]{@{}l@{}}USPTO-50K \end{tabular} \\
\midrule

New Drug Design & Bongini et al. (2021) \cite{MG2N2} & MG$^2$N$^2$ & GNN & \begin{tabular}[t]{@{}l@{}}Validity, Uniqueness, Novelty, \\VUN \end{tabular} & \begin{tabular}[t]{@{}l@{}}QM9, Zinc \end{tabular} \\

& Manu et al. (2024) \cite{GraphGANFed} & GraphGANFed & \begin{tabular}[t]{@{}l@{}}GCN \end{tabular} & \begin{tabular}[t]{@{}l@{}}Validity, Uniqueness, Novelty, \\Internal Diversity, QED, logP, \\SNN \end{tabular} & \begin{tabular}[t]{@{}l@{}}ESOL, QM8, QM9 \end{tabular} \\

& Macedo et al. (2024) \cite{MedGAN} & MedGAN & \begin{tabular}[t]{@{}l@{}}GCN \end{tabular} & \begin{tabular}[t]{@{}l@{}}Validity, Connected validity, \\ Novelty, Diversity \end{tabular} & \begin{tabular}[t]{@{}l@{}}PubChem, ZINC15 \end{tabular} \\
\bottomrule
\end{tabular}
\end{adjustbox}
\label{tab:litreview}
\end{table*}

\subsection{Metrics}

\subsubsection{Regression Metrics} Regression is to find a model describing how a dependent variable relates to independent variables, which can be used to predict the dependent variable for new or unseen data. Frequently used metrics for evaluating regression performance are listed below.\\
\begin{itemize}
    \item Mean Squared Error (MSE): Average squared distance between true and predicted values.
    \item Root Mean Squared Error (RMSE): Square root of MSE.
    \item Mean Absolute Error (MAE): Average absolute difference between true and predicted values.
    \item $R^2$: Proportion of variation.
    \item $r_m^2$: Modified $R^2$ to penalize deviation.
    \item Pearson Correlation (R): Linear correlation between true and predicted values.
    \item Spearman Correlation: Monotonic correlation between true and predicted values.
    \item Concordance Index (CI): Fraction of correctly predicted pairwise orderings.\\
\end{itemize}

\subsubsection{Classification Metrics} Classification models assign input data points to discrete, categorical outputs. Traditional metrics like accuracy can be misleading, especially in scenarios of imbalanced datasets. Thus, curve-based metrics are frequently used for classification as well. \\
\begin{itemize}
    \item Accuracy (ACC): Fraction of correct predictions.
    \item Precision (PREC, Positive Predictive Value, PPV): Correctly predicted positives out of all predicted positives.
    \item Recall (True Positive Rate, TPR, Sensitivity): Correctly predicted positives out of all actual positives.
    \item Weighted Precision/Recall: Precision/Recall weighted to account for class imbalance.
    \item F1-Score: Harmonic mean of precision and recall.
    \item Specificity (True Negative Rate, TNR): Correctly predicted negatives out of all actual negatives.
    \item Balanced Accuracy (BACC): Average sensitivity and specificity.
    \item Kappa (Cohen's Kappa): Agreement between classifiers.
    \item Matthews Correlation Coefficient (MCC): Balanced measure of binary classification performance.
    \item Top-k Exact Match Accuracy: Frequency of true labels among top-k predicted labels.
    \item Top-k Round-Trip Accuracy: Evaluates whether predicted reactants can produce the target product in forward synthesis.
    \item MaxFrag Accuracy: Evaluates whether the largest fragment in a predicted reaction sequence is correct.
    \item Ligand RMSD: Measure of how closely the predicted ligand pose matches the expected pose.
    \item Centroid Distance: Euclidean distance between the geometric center (centroid) of predicted and actual ligand conformations.\\
\end{itemize}

\begin{itemize}
    \item ROC-AUC (AUCROC): ROC: Receiver Operating Characteristic Curve; AUC: Area Under the Curve.
    \item PRC: Precision Recall Curve.
    \item AUPRC (AUPR, PRAUC): Area Under Precision Recall Curve.
    \item Adjusted logAUC: Log-scaled AUC to emphasize early enrichment.\\
\end{itemize}

\subsubsection{Generation Metrics} These are for drug generation.\\
\begin{itemize}
    \item Validity: Fraction of generated molecules that are chemically valid.
    \item Uniqueness: Fraction of valid molecules that are unique.
    \item Novelty: Fraction of valid molecules that are different from dataset examples.
    \item VUN: Fraction of valid, unique, and novel molecules out of the total generated.
    \item Internal Diversity: Estimation of chemical diversity in generated examples.
    \item Quantitative Estimation of Drug-Likeliness (QED): Probability that a molecule is a viable drug candidate.
    \item Octanol-water partition coefficient (LogP): Measure of molecule lipophilicity.
    \item Similarity to a Nearest Neighbor (SNN): Similarity of a generated molecule to its nearest neighbor molecule in the real dataset.
\end{itemize}

\subsection{Datasets}

The frequently used datasets in the papers surveyed, along with the links to access them, are described in Table \ref{tab:datasets}.

\begin{table*}[t]
    \centering
    \caption{Commonly Used Datasets for Drug Discovery}
    \label{tab:datasets}
    \renewcommand{\arraystretch}{1.3}
    \begin{tabular}{p{1.5cm} p{7cm} p{3cm} p{6cm}}
        \toprule
        \textbf{Dataset} & \textbf{Description} & \textbf{Number of Molecules} & \textbf{Reference} \\
        \midrule
        ESOL & Water solubility data for common organic small molecules. & 1128 & \url{https://moleculenet.org/datasets-1} \\
        FreeSolv & Experimental and calculated hydration free energies for small neutral molecules in water. & 642 & \url{https://moleculenet.org/datasets-1} \\
        Lipophilicity (Lipop) & Experimental results of lipophilicity (octanol/water distribution coefficient, logP) of molecules. & 4200 & \url{https://moleculenet.org/datasets-1} \\
        BBBP & Binary classification dataset of blood-brain barrier penetration. & 2053 & \url{https://moleculenet.org/datasets-1} \\
        BACE & Dataset of beta-secretase 1 (BACE-1) inhibitors. & 1513 & \url{https://moleculenet.org/datasets-1} \\
        ClinTox & Clinical toxicity dataset comparing FDA approved drugs and drugs that have failed in clinical trials due to toxicity. & 1491 & \url{https://moleculenet.org/datasets-1} \\
        SIDER & Recorded side effects and adverse reactions of marketed drugs. & 1427 & \url{https://moleculenet.org/datasets-1} \\
        Tox21 & Toxicity measurements of compounds across 12 targets. & 7831 & \url{https://moleculenet.org/datasets-1} \\
        ToxCast & Toxicity Forecaster, high-throughput toxicity screening of thousands of chemicals. & 8575 & \url{https://moleculenet.org/datasets-1} \\
        HIV & Molecules screened for their ability to inhibit HIV replication. & 41127 & \url{https://moleculenet.org/datasets-1} \\
        MUV & Maximum Unbiased Validation for virtual screening techniques. & 93087 & \url{https://moleculenet.org/datasets-1} \\
        DUD-E & Directory of Useful Decoys - Enhanced, contains active and decoy molecules for protein targets. & 22886 & \url{https://dude.docking.org/} \\
        PDBbind & Binding affinity data and 3D structural information for protein-ligand complexes. & 11908 & \url{https://moleculenet.org/datasets-1} \\
        ChEMBL & Large database of bioactive molecules with drug-like properties. & 2496335 & \url{https://www.ebi.ac.uk/chembl/} \\
        Davis & Binding affinities for 72 kinase inhibitors and 442 targets. & 68 & \url{https://www.kaggle.com/datasets/christang0002/davis-and-kiba} \\
        KIBA & Kinase Inhibitor BioActivity, drug-target bioactivity matrix for 52,498 chemical compounds and 467 kinase targets. & 2111 & \url{https://www.kaggle.com/datasets/christang0002/davis-and-kiba} \\
        Metz & Similar to Davis/KIBA, used for kinase inhibitor affinity prediction. & 1423 & \url{https://www.kaggle.com/datasets/christang0002/metz-dta} \\
        DrugBank & Comprehensive database for chemical, pharmacological, and pharmaceutical data on drugs and targets. & 18490 & \url{https://go.drugbank.com/} \\
        QM8 & Quantum Machine 8, dataset containing 20,000 synthetically feasible, small, and organic molecules with up to eight Carbon, Oxygen, Nitrogen, or Fluorine (CONF) atoms, representing quantum mechanical calculations of electronic fields and excited state energy. & 21786 & \url{https://moleculenet.org/datasets-1} \\
        QM9 & Quantum Machine 9, dataset containing 133,885 organic molecules composed of up to 9 CONF atoms with hydrogen for a maximum size of up to 29 atoms, with computed geometric, energetic, electronic, and thermodynamic properties reported. & 133885 & \url{https://moleculenet.org/datasets-1} \\
        MDAD & Microbe-Drug Association Database, supported by clinical and experimental evidence. & 1373 & \url{https://ieee-dataport.org/documents/mdadabiofilm} \\
        aBiofilm & Biological, chemical, and structural details of 5027 anti-biofilm agents. & 1720 & \url{https://ieee-dataport.org/documents/mdadabiofilm} \\
        DrugVirus & Known and predicted associations between antiviral drugs and viruses. & 231 & \url{https://drugvirus.info/} \\
        Fdataset & Drug–disease associations compiled from DrugBank and OMIM. & 593 & \url{https://zenodo.org/records/8357512} \\
        Cdataset & Expansion of Fdataset. & 663 & \url{https://zenodo.org/records/8357512} \\
        KEGG & Kyoto Encyclopedia of Genes and Genomes, database of pathways, genes, drugs, diseases, and compounds. & 12684 & \url{https://www.genome.jp/kegg/pathway.html} \\
        CTD & Comparative Toxicogenomics Database, resource linking chemicals, genes, phenotypes, and diseases. & N/A & \url{https://ctdbase.org/} \\
        STRING & Search Tool for the Retrieval of Interacting Genes/Proteins, protein-protein interaction network database. & N/A & \url{https://string-db.org/} \\
        UniProt & Universal Protein Resource, protein sequence and functional information. & N/A & \url{https://www.uniprot.org/} \\
        USPTO-50K & Dataset of 50K atom-mapped reactions from USPTO drug patents classified into 10 reaction classes. & N/A & \url{https://figshare.com/articles/dataset/USPTO-50K_raw_/25459573?file=45206101} \\
        \bottomrule
    \end{tabular}
\end{table*}

\clearpage


\section{Molecular Property Prediction}\label{sec:molecularproperty}
The first commonly explored area of GNNs in drug discovery is molecular property prediction, which can encompass the prediction of a broad range molecular properties such as quantum chemistry, physicochemical properties, bioactivity, and ADMET (Absorption, Distribution, Metabolism, Excretion, and Toxicity). In this category, nine recent papers were reviewed to provide a snapshot of the current state of GNNs for this application. 

The timeline of GNNs for molecular property prediction in this survey begins with a paper from 2019, in which Xiong et al. proposed the Attentive FP (fingerprint) model for molecular property prediction \cite{AttentiveFP}. Attentive FP is composed of stacked graph attention layers to extract both atom-level and molecule-level information, and a virtual supernode to aggregate atom features. The proposed model was compared to a variety of state-of-the-art models such as Message Passing Neural Network (MPNN), Neural FP, MultiTask, GC, Support Vector Machines (SVM), Random Forests (RF), Weave, ECFP with linear layer, and ECFP with neural net. The models were trained on two collections of datasets, the first of which for predicting various tasks such as solubility, malaria bioactivity, and photovoltaic efficiency. The second collection covered prediction in categories including physical chemistry (ESOL, FreeSolv, Lipop), bioactivity (MUV, HIV, BACE), and physiology and toxicity (BBBP, Tox21, ToxCast, SIDER, ClinTox). With respect to the previous best performance for each dataset, Attentive FP outperformed all but RF on BACE and SIDER. From these results, it is shown that Attentive FP is an effective framework for capturing molecular substructure patterns, for example hydrogen bonding and aromatic systems.

In 2020, Peng et al. proposed an enhanced GIN model, called MolGIN, specifically for application in ADMET prediction \cite{MolGIN}. While previous GIN models have ignored factors such as bond features and influence of neighboring nodes, MolGIN provides a solution to this by concatenating bond features and adjusting neighborhood weights using a gate unit. By incorporating these enhancements, this model is better suited to extract local structural information relevant to ADMET properties. The model was assessed both with regression datasets LD50, LogD$_{7.4}$, ESOL, and PPB in terms of RMSE and with classification datasets Ames, P-gp inhibitors, and Tox21 in terms of AUC. Compared to models such as RF, fully connected deep neural network (FDNN), GCN, and a baseline GIN, MolGIN showed improvements over all datasets for both regression and classification. The results also showed that a multi-task version of the model provided improvements compared to a single-task version over several datasets. Peng et al. conclude that although the model provided promising performance, it is difficult to identify which molecular substructures have the greatest importance in the prediction task, so future work will focus on improving this interpretability.  

In another contribution from 2020, Shen et al. introduced a model called MoLGNN for molecular property prediction based on chemical motif learning \cite{MoLGNN}. One of the most pressing challenges in the effective use of machine learning for drug discovery has been the lack of plentiful labeled data, and thus MoLGNN was proposed to improve performance when high quality data is unavailable. To address this, MoLGNN is pretrained on unlabeled data in a multi-task self-supervised learning setup, using components such as Graph Isomorphism Network Variational Auto-Encoder (GINVAE) for edge reconstruction, self-labeled motif learning, and supervised graph classification. Instead of focusing only on graph and node level information, MoLGNN also focuses on interpreting subgraph and chemical motif information, which provides meaningful insight into molecular properties and overall drug behavior. Model performance was assessed in terms of ROC-AUC on BACE, BBBP, SIDER, ClinTox, and HIV datasets, compared with the baseline ContextPred model with and without pretraining as well as ablations of the proposed model. Results show that the full MoLGNN model has many advantages and robust improvements in comparison with state-of-the-art methods, especially with reduced amounts of data. To test real-world applicability of the model, it was also tested using JAK1, JAK2, and JAK3 datasets that compile drugs for targeting Janus kinase, which was considered as a possible treatment for COVID-19. MoLGNN demonstrated impressive performance for each of these datasets in terms of ROC-AUC, showing the real-world utility of the model in full. In conclusion, the model proved to be an effective solution for sparse data with potential for novel applications, and future work could focus on determining more advanced GNN methods to add to the framework or replace current components for better performance.  

In 2021, the efficacy of traditional machine learning methods alongside GNN was explored by Deng et al. with the proposed hybrid model XGraphBoost \cite{XGraphBoost}. Typically, traditional machine learning models require manual feature engineering, while deep learning models automatically learn features from data. Thus, XGraphBoost makes use of this concept by leveraging GNN architectures such as a GCN, GGNN, and DMPNN to extract molecular features to be fed into an XGBoost model for the final output. This proposed model was benchmarked on datasets ESOL, FreeSolv, and Lipophilicity for regression evaluated with RMSE and datasets BACE, BBBP, Clintox, HIV, Tox21, ToxCast, and SIDER for classification evaluated with AUC. It was found that DMPNN in combination with XGBoost consistently produced the best results, and that the integration of XGBoost generally provided improvement beyond the performance of GNN alone. While this result is promising, the work is limited in the lack of comparison with other state-of-the-art methods of the time, the use of 1D SMILES representations within the XGraphBoost portion, and the limited number of samples in the datasets. Deng et al. suggest that the latter issue may be improved by integrating transfer learning or few-shot learning.  

The MolCLR model was introduced in 2022 by Wang et al. to consider contrastive learning of molecular representations \cite{MolCLR}. This model tackles the challenge of limited labeled data in using a self-supervised framework with contrastive learning, which maximizes the association between different orientations of the same molecules and minimizes the association between different molecules. The MolCLR model was implemented with a GCN and GIN as encoders alongside average pooling for feature extraction. The model was evaluated on the typical molecular property prediction datasets BBBP, Tox21, ClinTox, HIV, BACE, SIDER, and MUV for classification and FreeSolv, ESOL, Lipo, QM7, QM8, and QM9 for regression. In comparison with supervised RF, SVM, GCN, GIN, SchNet, MGCN, and D-MPNN. and N-Gram, MolCLR was shown to generalize better than many of the supervised methods. Wang et al. conclude that overall, MolCLR provides learned molecular representations that are transferrable and improve generalizability with limited data, though the model could be improved further by optimizing GNN backbones. 

In 2023, a mechanism-driven GNN model, aptly named MD-GNN, was proposed by Chen et al. \cite{MDGNN}. MD-GNN combines GCN and a Graph Attention Network (GAT) in message passing layers, uses feature fusion layers to combine graph embeddings with descriptors, and combines outputs with mechanism-driven theoretical predictions based on weighted averaging in a correction block. The model was tested on a dataset collected from PubChem alongside other models such as ene-s2s, GAT, GraphSage, SchNet, Linear Regression, RF, Gradient Boosting, and ANN (multi-layer perceptron), outperforming all of them in terms of MAE. The correction block was also added to feature fusion supplemented models such as GNN, ene-s2s, GAT, GraphSage, and SchNet to successfully show that the correction block is a valuable addition to any model. MD-GNN was also used to propose nine candidate drug molecules related to ibuprofen, demonstrating the potential of the model in drug discovery.

The first contribution from 2024 to be reviewed in this category was an exploration of transfer learning in GNN for molecular property prediction, conducted by Buterez et al. \cite{Buterez}. The main motivation for this model was driven by the lack of abundant and accessible high-fidelity data, creating a goal to improve molecular property prediction in the multi-fidelity setting. To do this, low-fidelity data is leveraged to improve prediction on high-fidelity targets. Buterez et al. proposed a workflow that replaces mean or sum-based pooling with neural readouts to improve transfer learning while also making use of Variational Graph Autoencoders (VGAEs) for embedding-based transfer. The proposed methods were evaluated on datasets from AstraZeneca and PubChem alongside QMugs and QM7b, which showed superior performance of neural readouts compared to mean and sum pooling in terms of MAE. The results confirm that neural readouts and embedding-based transfer learning on low-fidelity data surpasses state-of-the-art methods and improves generalization without the need for high-fidelity training data. This reinforces the potential for transfer learning in GNN for molecular property prediction and provides insight into the optimization of transfer learning methods. 

Another model focused specifically on ADMET prediction was proposed by Xu et al. in 2024, named ChemXTree \cite{ChemXTree}. This model is another hybrid framework, combining an MPNN based encoder with a Gate Modulation Feature Unit (GMFU) for optimized feature selection and a Neural Decision Tree (NDT) for classification output. ChemXTree was compared to many state-of-the-art models such as D-MPNN, Attentive FP, N-GramRF, N-GramXGB, PretrainGNN, GROVERbase, GROVERlarge, GraphMVP, MolCLR, GEM, Uni-Mol, Graphormer, AutoML, InfoGraph, GROVER, MAT, GAT, and XGBoost on many datasets such as BBBP, BACE, HIV, ClinTox, AMES, CYP3A4\_Substrate, CYP3A4\_inhibitor, CYP2D6\_Substrate, CYP2D6\_inhibitor, CYP2C9\_Substrate, CYP2C9\_inhibitor, and Bioavailability. The results revealed that ChemXTree generally outperformed or provided comparable performance to the other models across the various datasets. Ablation studies also showed the importance of the GMFU component in the model’s performance, and that the NDT component was more effective than alternatives such as FFN and XGBoost. Some directions Xu et al. suggest for future work include model pretraining, optimizing computational complexity, simplifying the workflow, and reducing the number of parameters. 

In 2025, Satheeskumar introduced an ensemble approach specifically for ADME (Absorption, Distribution, Metabolism, and Excrection) prediction, referred to as a Stacking Ensemble \cite{StackingEnsemble}. Based on the different strengths of various learning models, this ensemble approach combines GNN, Transformers, and traditional models such as Random Forest and XGBoost to create a robust model that provides accurate predictions. Compared to individual models such as GNN, Transformers, Random Forest, XGBoost, Neural Networks, Support Vector Regression (SVR), and Lasso Regression, the Stacking Ensemble provided the best performance in terms of R$^2$, MAE, and RMSE on a dataset from ChEMBL. Pearson and Spearman coefficients were also used to show the correlation between actual pharmacokinetic parameter values and those predicted by the model, which demonstrated high predictive accuracy. A few case studies were also conducted based on specific molecules with high lipophilicity, complex chemical structure, and other extreme pharmacokinetic properties, which showed the robustness and generalizability of the Stacking Ensemble in real-world scenarios. Satheeskumar indicates that the model has room for improvement in terms of limited data and interpretability, and directions for future work include the incorporation of more real-world and clinical data, employment of transfer learning to reduce computational burden and federated learning to preserve data privacy, and improvement of interpretability to increase trust in the pharmaceutical industry.    

\section{Drug-Target Binding Affinity Study}\label{sec:dta}

The second reviewed category that has been widely explored with GNN is drug-target binding affinity, which is also commonly referred to as protein-ligand binding affinity. Although this category could also technically fall under molecular property prediction, it has been considered as its own thread due to the focused attention it has received in experimental work. Literature on protein pocket prediction and docking has also been considered in this category due to its relevance in the problem of binding affinity prediction. A total of eight papers have been reviewed in this category with an attempt to thoroughly capture the current state-of-the-art.  

In 2019, Lim et al. introduced a novel GNN model for predicting drug-target interaction by incorporating a distance-aware graph attention algorithm alongside directly extracted 3D structural information from the protein-ligand binding pose \cite{Lim}. Before this, traditional docking methods were used to predict drug-target binding affinity, and more recently deep learning methods have been explored, both of which have provided a baseline for comparison in this study. The proposed model was trained and tested on the DUD-E and PDBbind datasets for virtual screening and binding pose prediction respectively, as well as the ChEMBL, MUV, and IBScreen libraries. Metrics such as AUROC, adjusted Log-AUC, PRAUC, sensitivity, specificity, and balanced accuracy were used to assess performance, and the proposed model outperformed previous methods in all metrics for each dataset except for MUV, where a previous graph convolutional neural network showed better performance. Overall, the performance of this model on DUD-E and PDBbind was quite promising with AUROC scores of 0.968 and 0.935 respectively, though this performance dropped with ChEMBL and MUV with respective AUROC values of 0.633 and 0.536. These results suggest that although the model is promising, it is limited in generalization. Lim et al. conclude by suggesting that Bayesian neural networks could be used in uncertainty quantification of drug-target interaction predictions, since sufficient quantities of high-quality data are not always available.  

Nguyen et al. proposed the GraphDTA model in 2021 \cite{GraphDTA}, which would go on to be the baseline for much of the successive work on this thread. In this model, drug molecules are represented as graphs to be interpreted by a GNN, and proteins are represented as 1D strings to be interpreted by a Convolutional Neural Network (CNN). In order to develop the most effective model, several candidate GNN architectures, including GCN, GAT, GIN, and GAT-GIN were tested and compared on the Davis and KIBA datasets alongside previous deep learning models for drug-target binding affinity prediction. It was found that the GIN and GAT-GIN variants produced the best performance with Davis and KIBA respectively, outperforming the baseline models in both cases with respect to metrics such as CI and MSE. When analyzing the GIN prediction error for specific drugs in each dataset, it was found that several drugs seemed to contribute disproportionately to overall error, which provides further insight into the performance and potential of the model. In conclusion, Nguyen et al. suggest that the model could further be improved by also representing proteins as graphs, as this method and all those prior have only considered proteins as 1D strings that omit structural information.  

In 2022, MGraphDTA was proposed by Yang et al. to expand upon GraphDTA \cite{MGraphDTA}. MGraphDTA presents a multiscale GNN (MGNN) to capture both local and global molecular structures alongside a multiscale CNN (MCNN) to capture both domain and motif-level information of target proteins. In addition, Yang et al. also introduced a novel explanation method based on Gradient-weighted Affinity Activation Mapping (Grad-AAM) to improve interpretability. The model was trained and tested on Human and C. elegans datasets for classification as well as Davis, Filtered Davis, KIBA, Metz, and ToxCast datasets for regression. First, both MGNN-CNN and MGNN-MCNN models were compared with baseline models on Human and C. elegans in terms of precision, recall, and AUC, which revealed that the MCNN component was beneficial for performance, though both proposed models outperformed the baselines. This superior performance held up across each dataset in terms of MSE, CI, $r_m^2$ index, and Spearman correlation. The Grad-AAM visualization was also found to be helpful in visualizing important chemical substructures as well as structural alerts related to toxicity, which proved to be more accurate than those derived from GAT. Though this model provided valuable advances in binding affinity prediction, it is limited in that it was only tested on relatively small datasets, so a direction for future work would be to test it with larger datasets to ensure generalization. Another improvement identified by Yang et al. would be to focus more on binding mode visualization with Grad-AAM to better understand structural alerts and toxicity.  

The model GraphscoreDTA was introduced by Wang et al. in 2023 as another optimized GNN for protein-ligand binding affinity \cite{GraphScoreDTA}. This model presents a unique architecture that includes a Bitransport Information Mechanism for exchanging information between protein and ligand representations, a multi-input GNN for proteins, ligands, and pocket-ligand interactions, multihead attention, Gated Recurrent Units (GRUs), and skip connections. GraphscoreDTA also integrates AutoDock Vina distance terms to score binding affinity using physical concepts. A dataset from PDBbind was used to train the model while CASF2016 and CASF2013 datasets were used for testing. The proposed GraphscoreDTA model was compared with baselines such as DeepDTA, DeepDTAF, Pafnucy, BAPA, and a fusion model, with GraphscoreDTA outperforming each across the board in terms of R, CI, RMSE, MAE, and SD. Several other experiments were conducted on data such as AlphaFold structures, cyclin-dependent kinases (CDKs), target selectivity, and homologous protein families, which further showed the utility of the model in terms of generalization, ranking of top-affinity inhibitors, drug-target selectivity prediction, and specificity. Wang et al. also discuss that the size of the protein pockets as well as the molecules has an impact on binding affinity, and thus future work will explore the integration of the size information to improve predictions. 

Another emergent model from 2023 was proposed by He et al., named NHGNN-DTA \cite{NHGNNDTA}. This model, based on a node-adaptive hybrid GNN, combines drug molecules with target proteins in a hybrid graph, allowing for greater interpretation of the interactions between drug and target subgraphs. The model also integrates BiLSTM and multihead attention to extract features for both drug graphs and protein sequences, and implements multilayer GIN to predict affinity from the hybrid graph. The model was evaluated on the Davis and KIBA datasets and compared to models such as DeepDTA, MT-DTI, GraphDTA, rzMLP, EnsembleDLM, FusionDTA, and MGraphDTA. In terms of MSE, CI, and $r_m^2$, NHGNN-DTA outperformed all other models across the board, even when evaluated on cold drug and cold target scenarios. A case study was also conducted for NHGNN-DTA to screen FDA-approved drugs for application in treating SARS-CoV-2 Omicron variants, in which it identified Amyl Nitrate as a candidate, which shows a promising affinity for the target.  

Another hybrid GNN model, graphLambda, was introduced by Mqawass and Popov in 2024 \cite{graphLambda}. The graphLambda model is considered a hybrid model in that it combines three GNN architectures in one model: GCN, GAT, and GIN. With this, graphLambda leverages the strengths of each GNN architecture, such as local feature extraction with GCN, inference of important atoms from GAT, and representational power from GIN. A subset of PDBbind was used to train the model, and structure-aware clustering was used in data splitting to avoid overfitting. CASF16 and CSAR HiQ NRC datasets were used to evaluate the model alongside graphDelta, CNN-GNN Fusion, RF-Score, and Kdeep, where graphLambda achieved state-of-the-art performance, though there were variations in performance depending on the splitting method used. An ablation study also showed that every component is important to the model’s performance, so the model is best applied in its entire proposed form. GraphLambda is a robust model that surpasses previous models on various train-validation splits and data configurations. 

Another contribution from 2024 is the SSR-DTA model proposed by Liu et al. \cite{SSRDTA}. This model introduces a substructure-aware multi-layer GNN that implements GAT for drugs as well as a Bi-graph Neural Network (BiGNN) to interpret sequences and graph-based features of proteins simultaneously. Benchmark datasets Davis, KIBA, Metz, and BindingDB were used to evaluate the model in comparison with other models such as KronRLS, SimBoost, DeepDTA, DeepCDA, GraphDTA-GIN, MGraphDTA, WGNNDTA, FusionDTA, BiCompDTA, GLCN-DTA, and 3DProtDTA. In terms of MSE, CI, Pearson, and $r_m^2$, the full SSR-DTA model outperformed each model on every dataset. SSR-DTA was also compared with GraphDTA and MGraphDTA in cold start experiments, where it also outperformed both. In conclusion, SSR-DTA is a robust and interpretable model that generalizes well, although it is still limited in performance on different distributions of training and test sets, which could be a direction for future work.  

The most recent contribution for this category is the model PPDock \cite{PPDock}, proposed by Du et al. in 2025. Although this model more specifically deals with prediction of drug molecule docking conformation in a specific target protein pocket, this concept directly relates to binding affinity and drug-target interaction, so it has been considered as a recent contribution to this category. In two stages, this model predicts both protein pocket prediction and docking within the predicted pocket. The model also necessarily considers both the drug and the target as graphs, using Equivariant GNN (EGNN) and cross-attention encoding to consider interactions between the drug and target pair. The dataset used to train and evaluate the model was collected from PDBbind, and the performance of the model was compared with other docking methods such as QVina-W, GNINA, Smina, Glide, Vina, EquiBind, TANKBind, E3Bind, DiffDock, and FABind. PPDock outperformed most other docking methods in terms of Ligand RMSD and Centroid Distance, except for a few where it placed second. Additionally, PPDock was tested on unseen proteins, showing the best performance in that case as well. Du et al. conclude that PPDock is accurate and efficient with generalization capability, though they are still committed to improving it. 

\section{Drug-Drug Interaction and Synergy Prediction}\label{sec:ddi}

The next reviewed category evaluates drug behavior beyond the target, specifically focusing on the prediction of drug-drug interactions and synergistic effects. These factors are essential in drug discovery as they can significantly impact drug efficacy and safety, especially in the presence of complex diseases and comorbidities. In this section, four relevant GNN models are reviewed.

In 2022, Wang et al. proposed the DeepDDS model to predict synergistic drug combinations for cancer treatment \cite{DeepDDS}. DeepDDS takes two drugs as input to a GNN, tested with both GAT and GCN, and a cancer cell line as input to an MLP. The outputs are then concatenated and passed through fully connected layers to predict synergistic effect. The drug molecule dataset was collected from DrugBank, and the cancer cell line dataset was taken from CCLE. The performance of the model was compared to other models such as XGBoost, RF, GBM, Adaboost, MLP, SVM, AuDNNSynergy, TranSynergy, DTF, and DeepSynergy, and both the GAT and GCN variants of DeepDDS outperformed the others in terms of ROC AUC, PR AUC, ACC, BACC, PREC, TPR, and KAPPA. The performance of the two variants was nearly identical, though GAT was found to be superior albeit by a narrow margin. The study also performed leave-one-out experiments, with DeepDDS-GAT still prevailing. The models were also tested on an independent test set from AstraZeneca, and DeepDDS-GCN generally had the better performance in this case. DeepDDS was also applied to predict novel synergistic drug combinations for the A375 melanoma cell line, and at least five predicted combinations were supported by previous literature. While the performance of this model is exceptional, Wang et al. discuss the limitation that it was trained on very little data, which may have attributed to any shortcomings. Future work could focus on training and testing the model with more data and potentially integrating knowledge graphs or self-supervised learning.  

Zhang et al. proposed a GNN to predict drug-drug interaction in 2023, named EmerGNN \cite{EmerGNN}. One issue with emerging drugs is that their interactions are typically unknown, which causes issues in using computational methods that require large amounts of drug-drug interaction data. To address this, EmerGNN instead leverages biomedical networks that provide insights into the connections between drugs. EmerGNN focuses on the subgraphs of paths between drug pairs as well as the propagation of information from one drug to another using a flow-based GNN. Datasets including DrugBank and TWOSIDES were used to evaluate the model alongside other methods and models such as MLP, Similarity, CSMDDI, STNN-DDI, HIN-DDI, MSTE, KG-DDI, CompGCN, Decagon, KGNN, SumGNN, and DeepLGF. In terms of metrics such as F1-Score, accuracy, and Kappa for DrugBank and PR-AUC, ROC-AUC, and accuracy for TWOSIDES, EmerGNN outperformed all models across the board. Although these results are quite promising, EmerGNN is still limited by what is or isn’t in the biomedical network, much like the original issue with other computational methods for drug-drug interaction not having access to enough information. Still, this model showed better performance than previous methods, and other future work could involve testing the model in other scenarios such as drug-target, protein-protein, or disease-gene interactions.  

In 2024, Liu et al. proposed SynerGNet, another GNN model for anticancer drug synergy assessment \cite{SynerGNet}. SynerGNet integrates heterogeneous biological features into a protein-protein interaction network to create cancer-specific feature graphs. The model takes the reduced graph of a pair of drugs and a cancer cell line as input to a GCN that generates embeddings over two modules, then aggregates the embeddings with a jumping knowledge network (JK-Net). The model was trained on the AZ-DREAM Challenges dataset and tested on the DrugCombDB dataset, and the performance was compared to other methods such as RF and PRODeepSyn, each with and without augmented data. In terms of AUC, BAC, PPV, FPR, and MCC, SynerGNet showed strong performance and outperformed baselines on the test set, and the performance of all models was improved with augmented data. While these results are promising, Liu et al. identify that the model is limited when it comes to full-sized graph data, as the graphs are reduced for use with this model, and the improved performance with augmented data may indicate potential limitations in scenarios where augmented data is not available.  

In 2025, the GAINET model was introduced by Das et al. for drug-drug interaction prediction \cite{GAINET}. GAINET, a Graph Attention Interaction Network, combines GAT with a co-attention mechanism that determines the relationship between elements in two different datasets, and knowledge graphs are also constructed to model the interactions between drug pairs. Using examples from the DrugBank database, the model was assessed in comparison with other baselines from literature such as MR-GNN, NeoDTI, SHGCL-DTI, MDNN, DDIMDL, and KGNN in terms of accuracy, AUPR, F1-Score, precision, and recall. GAINET consistently outperformed the baselines while also providing interpretability through visualizations of drug-drug interaction predictions with relevant substructures highlighted. In order to further improve performance, Das et al. suggest that future work will focus on integrating larger datasets and additional biological networks, as well as expanding predictive performance to multi-task learning. 

\section{Microbiome Interaction Study}\label{sec:microbiome}

Continuing the theme of drug interactions beyond the target, this section explores the application of GNNs in the prediction of human microbiome interaction, which encompasses microbe-drug associations and drug susceptibility in the human microbiome. The human microbiome plays an integral role in how medications are processed in the body, as can several personal factors beyond the target, and thus it is important to consider this factor within the drug discovery process. Here, four papers of this theme are reviewed. 

In 2022, Yang et al. proposed a GNN model to infer microbe-drug associations, named MKGCN to indicate that it blends multiple kernel fusion with GCN \cite{MKGCN}. In this framework, a heterogeneous graph of microbes and drugs is constructed for a GCN to extract multi-layer features, and the embeddings are then used to calculate kernel matrices. The multiple kernel matrices are fused, and Dual Laplacian Regularized Least Squares is used to determine microbe-drug associations from the combined kernel. Three commonly used microbe-drug association datasets, MDAD, aBiofilm, and DrugVirus, were used, and the performance of the model was compared to models: KATZHMDA, NTSHMDA, WMGHMDA, IMCMDA, GCMDR, BLM-NII, WNN-GIP, and GCNMDA. MKGCN outperformed all baselines on all datasets in terms of AUC, and almost all in terms of AUPR. A case study also showed that the model is capable of effectively predicting SARS-CoV-2-associated drugs, with nine out of the top 10 predictions being supported by literature.  

In 2024, GCGACNN was proposed by Su et al. as another GNN solution for microbe-drug association prediction \cite{GCGACNN}. GCGACNN is a hybrid model combining GCN, GAT, CNN, and RF for the final classification. The model analyzes the relationships between microbes, drugs, and diseases with matrices for microbe-drug, microbe-disease, and drug-disease associations. The data was taken from the MDAD dataset as well as literature sources, and the model was evaluated alongside existing methods such as HMDAKATZ, GCNMDA, EGATMDA, Graph2MDA, and GACNNMDA. GCGACNN showed outstanding performance in terms of AUC, AUPR, and F1 score, though it did not outperform baselines in terms of accuracy. Su et al. identify that the model is limited by sparse data structure and suggest that the model could be enhanced by considering additional biological data. 

In another contribution from 2024, Maryam et al. proposed a new method to explore human microbiome interaction with GNNs \cite{Maryam}. The human gut microbiome plays an integral role in how medications are processed in the body, and thus it is important to consider this factor within the drug discovery process. Maryam et al. proposed GNN models such as GCN, GAT, and GINCOV to classify drug susceptibility to depletion in gut microbiota, and gathered SMILES data from related literature to test the models. It was found that GINCOV had the best performance of the three in terms of accuracy, sensitivity, specificity, balanced accuracy, and weighted precision and recall, which also outperformed a previously proposed method, the Extra Trees algorithm. Overall, this proposed model is promising for microbiome interaction prediction in the early stages of drug discovery as well as personalized medicine, although the performance may be improved with larger datasets. 

Most recently in 2025, Wang et al. proposed another GNN model for microbiome interaction, called GutMDA \cite{GutMDA}. This model uses multiple GCNs to predict drug-microbiome associations, and combines microbiota-microbiota similarity, drug-drug similarity, and disease-disease similarity in a triple network. Datasets MASI, MDAD, aBiofilm, and DrugVirus were used to test the model, and models such as Graph2MDA, GCNMDA, SCSMDA, MKGCN, JDASA-MRD, and HKFGCN were used for comparison. GutMDA outperformed all models in terms of AUPR, though did not outperform HKFGCN in terms of AUC. An ablation study showed that each part of the triple network is important for performance, especially the integration of disease-disease similarity. The results show that GutMDA is an effective model for predicting drug-microbiome interactions and associations, with a significant fraction of predicted associations being validated in PubMed literature. There are a few limitations to this model however, including the fact that it cannot tell whether drugs are affecting microbiota or if microbiota is affecting the drugs, and there is also a lack of diverse and quality data that would improve model performance.  

\section{Drug Repositioning}\label{sec:repositioning}

While the previous categories in this survey focused on the prediction of properties and interactions, this next category covers the use of GNNs for drug repositioning, also referred to as drug repurposing. This term explores the application of existing approved drugs for the treatment of other targets, which has the potential to accelerate the drug discovery process and provide solutions for novel diseases and epidemics as they emerge. This section covers five proposed GNN models for drug repositioning.  

In 2021, Li and Pan introduced a hybrid similarity side information graph neural network, or HSSIGNN, as a computational drug repositioning model \cite{HSSIGNN}. HSSIGNN pairs the GNN ability to learn hidden feature representations of drugs and diseases with the concept of side information from recommender systems to predict the probability of a drug being an effective treatment for a disease. The model takes a drug-disease association matrix, its transpose, a drug similarity matrix, and a disease similarity matrix as input after dimensionality reduction to extract hidden features along with similarity side information. Fdataset and Cdataset were used to compare the performance of HSSIGNN to traditional machine learning classifiers such as Linear Regression, Random Forest, and Decision Tree, with HSSIGNN significantly outperforming in terms of AUC and AUPR. Li and Pan conclude that the model is effective in its learning ability with better generalizability than machine learning benchmarks, showing the superiority of GNNs for this application. 

In 2022, a computational drug repurposing model was introduced by Doshi and Chepuri, called GDRnet \cite{GDRnet}. The model architecture consists of an encoder based on the Scalable Inception Graph Neural Network (SIGN) model and a decoder that predicts the likelihood of drug-disease treatment based on a scoring function. In this work, a drug repurposing network is devised using a four-layer heterogeneous graph connecting drugs, diseases, target genes, and affected anatomies, thus posing drug repurposing as a link prediction problem. The graph is constructed from the drug repurposing knowledge graph (DRKG) that contains data from databases such as DrugBank, Hetionet, GNBR, STRING, IntAct, and DGIdb. The performance of the model was compared to Graph-SAGE, GCN, GAT, HINGRL, and Bipartite-GCN in terms of AUROC and AUPRC, and GDRnet demonstrated superior performance in link prediction and drug ranking. An ablation study showed that the model is most effective with all four layers, showing a major improvement when genes are included and a minor improvement when anatomies are included. A case study was also conducted to predict drugs for COVID-19 treatment based on similar diseases, which yielded several drugs such as ivermectin, sirolimus, and dexamethasone. Doshi and Chepuri conclude that the work can be extended in several directions, and suggest that the model predictions may be improved by including information such as drug side effects, patient comorbidities, and drug interactions for synergistic effects. 

In another contribution from 2022, Gu et al. proposed the REDDA model, short for Relations-Enhanced Drug-Disease Association prediction \cite{REDDA}. REDDA incorporates three attention blocks with a heterogeneous GCN backbone to handle large-scale data with five biological entities (drugs, proteins, genes, pathways, and diseases) and 10 relations (drug-drug, drug-protein, protein-protein, gene-gene, gene-pathway, pathway-pathway, pathway-disease, disease-disease, and drug-disease) for the goal of drug repositioning. The extensive drug-disease association benchmark was constructed from Fdataset, Cdataset, KEGG, DrugBank, CTD, STRING, and UniProt. Compared to methods such as SCMFDD, MBiRW, DDA-SKF, NIMGCN, HINGRL-Node2Vec-RF, HINGRL-DeepWalk-RF, LAGCN, and DRWBNCF on the proposed benchmark dataset, REDDA outperformed all in terms of AUC, AUPR, F1-Score, accuracy, recall, and specificity, with second place on precision. REDDA was also tested on B-dataset against NIMCGCN, HINGRL-DeepWalk-RF, LAGCN, and DRWBNCF, in which case it outperformed all across the board. Ablation studies also showed that removing any components such as biological relations, topological subnet, or attention mechanisms significantly reduced performance. In a case study, REDDA was able to identify new indications and therapies, which demonstrates the knowledge-aware prediction of the model. With this, Gu et al. suppose that future work should focus on incorporating more biological associations and larger datasets in addition to integrating a more advanced backbone in the framework. 

In 2023, the AntiViralDL model was proposed by Zhang et al. for computational antiviral drug repurposing based on virus-drug association \cite{AntiViralDL}. AntiViralDL make use of LightGCN, an architecture that simplifies graph convolution into lightweight neighborhood propagation, as well as contrastive learning to enhance representational capacity and handle data sparsity. DrugVirus2 and US FDA datasets were used to construct a reliable virus-drug association bipartite graph for the model to derive embedding representations of viruses and drugs. With this merged dataset, AntiViralDL outperformed baseline models VDA-DLCMNMF, DRRS, IRNMF, and VDA-KATZ in terms of AUC and AUPR. The real-world applicability of AntiViralDL was assessed with a case study on COVID-19 drug repurposing, where the model predicted approved drugs such as Baloxavir marboxil, Laninamivir octanoate, Peramivir, Arbidol, and Oseltamivir, many of which are supported by literature or clinical trials. Zhang et al. conclude that AntiViralDL could be further improved in terms of generalization capability and predictive accuracy through future work with transfer learning and integration of multi-source heterogeneous features of viruses and antiviral drugs. 

In 2025, a Multi-view Representation Learning method with Heterogeneous Graph Neural Network, MRLHGNN, was introduced by Peng et al. for drug repositioning \cite{MRLHGNN}. In this framework, a heterogeneous graph is constructed using data from Fdataset, Cdataset, KEGG, CTD, DrugBank, STRING, and UniProt to include entities such as drugs, diseases, proteins, and pathways as well as their associations. View-specific graphs are introduced using meta-paths to represent local semantic relationships between nodes, and a transformer mechanism is used to aggregate semantic embeddings across views. An auto multi-view fusion decoder with attention mechanism is also implemented to predict drug-disease associations. When compared to baselines such as NTSIM, BNNR, NIMCGCN, HGIMC, LAGCN, MilGNet, DRWBNCF, REDDA, and DRHGCN, MRLHGNN generally surpassed the performance of the others on metrics AUC, AUPR, F1-Score, accuracy, recall, specificity, and precision, with a few exceptions. An ablation study was conducted to highlight the strength of the full model, as the model performed worse with components removed. Case studies were also conducted on non-small cell lung cancer and piroxicam drug to demonstrate the predictive capabilities of MRLHGNN from the perspective of both diseases and drugs, which produced 11 literature supported predictions out of 15 in both cases. Overall, MRLHGNN is a robust and effective model that provides interpretability, and some directions for future work include addressing class imbalance and exploring supervised contrastive learning as well as hypergraph learning. 

\section{Retrosynthesis}\label{sec:retrosynthesis}

Another essential category in the process of drug discovery is retrosynthesis and reaction prediction, which deals with the identification of synthetic routes for candidate drug molecules by reverse engineering from product to reactants. A necessary preface for this section is the distinction between retrosynthesis planning models that are template-based, template-free, and semi-template-based. Template-based models, as the name suggests, use templates that define known reactive rules for molecular changes during reactions, while template-free methods directly transform products into reactants. Semi-template-based models, on the other hand, predict final reactants through intermediates generated in multiple steps. In this section, a total of five relevant papers were reviewed: one template-based model followed by four semi-template-based models.  

In 2021, Chen and Jung introduced a graph-based retrosynthesis framework called LocalRetro, named for its focus on local molecular structures \cite{LocalRetro}. The motivation for this work originates from the intuition that molecular changes will mainly occur on a local scale during chemical reactions, yet traditional retrosynthesis models have neglected local information with the use of SMILES or fingerprints, which only consider the global molecular structure along with information that may be irrelevant. By using local reaction templates such as atom reaction templates, bond reaction templates, and multiple change reaction templates, LocalRetro is able to focus on relevant local structures. To account for nonlocal reactions and relationships between atoms and bonds, the model also includes Global Reactivity Attention (GRA) with multihead self-attention, differentiated from the typical local attention of GAT. Compared to other retrosynthesis models such as GLN, G2G, GraphRetro, Augmented Transformer, and MEGAN as well as a version of LocalRetro without GRA on the USPTO-50K dataset, the full model provided the best performance in terms of top 3, 5, 10, and 50 exact match accuracy and top 1, 3, 5, 10, and 50 round-trip accuracy. LocalRetro and LocalRetro without GRA were also compared using the USPTO-MIT dataset in terms of top-k exact match and round-trip accuracy, further proving the superior performance of the full model, the importance of GRA, and the scalability of the model to larger and noisier datasets. The model was also shown to correctly predict synthetic routes for drug candidates such as lenalidomide, salmeterol, a 5HT6 receptor ligand, and two DDR1 kinase inhibitors. With the promising performance of LocalRetro, Chen and Jung suggest that future work should involve larger datasets that include previously unconsidered information on reaction conditions such as reagents, temperature, and pH values.  

Han et al. proposed the GNN-Retro model in 2022 to explore synthetic route cost estimation, which has been proven to effectively streamline the otherwise vast search space \cite{GNNRetro}. Essentially, the estimated cost of one reaction is composed of the costs of each reactant, and the cost of each reactant is determined by its atoms and structure. Using Morgan fingerprints, which uniquely represent each molecule as a binary vector, it is assumed that molecules with similar fingerprints have similar cost. In this work, two versions of GNN-Retro are proposed: one that uses cosine similarity thresholding on fingerprints to connect molecules with similarities as weights, and the other where embeddings are learned from fingerprints to construct a k-NN graph from cosine similarities and embeddings. For a fair comparison with existing methods, A* search, an algorithm that finds the shortest path between nodes on a graph, was used to generate the final synthetic route. Both versions of GNN-Retro were compared to other cost estimation methods such as Greedy Depth-First Search (DFS), Monte Carlo Tree Search (MCTS), Depth-First Proof-Number Search with Heuristic Edge Cost (DFPN-E), Retro*-0, and Retro* using the USPTO dataset, which showed that both versions had a higher success rate than previous methods with the threshold method being the overall best. Ablation studies were also conducted to show the value of including the GNN component as well as partial ordering loss, as the full model provided significantly better performance. Han et al. conclude that GNN-Retro provides an effective solution for estimating the synthetic cost of molecules in retrosynthesis planning, which handles data sparsity and outperforms all existing methods studied. 

In the next reviewed paper from 2022, Zhang et al. introduced a chemistry-informed molecular graph, or CIMG, for use with a series of GNN models for retrosynthesis planning \cite{CIMG}. Based on the intuition that properties such as NMR chemical shifts, bond dissociation energies, and reaction conditions including solvents and catalysts are integral to chemical reactions, CIMG was created to encode this information as vertex features, edge features, and global features respectively. Though the main contribution of this paper is the CIMG, Zhang et al. also proposed a retrosynthesis planning process comprised of four sequential GNN models for reaction template selection, catalyst and solvent prediction, and reaction plausibility determination. The template selection model, which takes a CIMG descriptor as input, is comprised of six MPNN layers followed by a regular neural network that outputs a vector of reaction template probabilities, which outperformed the accuracy of other models such as highway-NN and a basic GNN model with only general atomic properties as input, showing the significant improvement of using the proposed chemical properties. The catalyst and solvent prediction models take a reactant CIMG vector, a product CIMG vector, and a one-hot reaction template vector as inputs to generate a probability vector for each catalyst or solvent in the reaction. In a comparison with versions of these models that do not include the reaction template vector input, it was shown that the performance of the models is significantly improved by including the reaction template information. Finally, the reaction plausibility determination GNN model receives the information for the reactants, product, catalyst, and solvent, and produces a probability value between 0 and 1 to assess the plausibility of the reaction. This model was able to achieve high plausibility scores and impressive ROC, while a version of the model without catalyst/solvent information performed much worse. Overall, this study presented a new perspective on retrosynthesis prediction with GNNs, and Zhang et al. conclude that even more relevant reaction information such as temperature, pressure, mixing methods, and reaction kinetics could be integrated in future work.  

A contribution from 2023 came from Zhong et al. with the proposed model Graph2Edits, based on the concept that reactions can be represented as a series of molecular graph edits \cite{Graph2Edits}. In this work, a variety of edit types are defined, such as bond deletion, bond change, atom change, leaving group attachment, and termination (end of edit sequence). Graph2Edits uses an autoregressive model to predict a sequence of edits based on outputs from a D-MPNN encoder that sequentially takes the product graph, intermediate graphs, and reactant graphs as inputs to generates atom and bond features, graph representations of atom and bond edits, and termination symbol predictions. The USPTO-50K dataset was used to compare the model to previous methods including template-based Retrosim, Neuralsym, GLN, and LocalRetro, template-free SCROP, Augmented Transformer, GTA, Graph2SMILES, and Dual-TF, and semi-template-based G2G, RetroXpert, RetroPrime, MEGAN, and GraphRetro in terms of top-k exact match accuracy. The model was also compared to LocalRetro, MEGAN, and GraphRetro in terms of top-k round-trip accuracy and MaxFrag accuracy. The results showed that Graph2Edits provided improved or comparable performance to the state-of-the-art models in many cases, which demonstrated its robustness and generalization. Despite this, its limitations include a lack of information on reaction conditions, bias from incorrect mappings between product and reactants, and an inability to attach the same leaving group to more than one atom in a graph, all of which will need to be addressed in the future work. 

In 2025, Ye et al. improved upon previous work with the proposed model State2Edits, which defines edits in terms of their state \cite{State2Edits}. In this framework, the main state contains edits such as atom edit, bond edit, motif edit, transform, and terminate, while the generate state only contains generate edit and terminate. Using a D-MPNN encoder, State2Edits predicts a sequence of edits to generate probable reactants for a target product. The model was compared with template-based methods such as GLN, LocalRetro, RetroComposer, RetroDCVAE, and RetroKNN, template-free methods including SCROP, Graph2SMILES, NAG2G, and PMSR, and semi-template-based methods e.g. G2G, GTA, GraphRetro, MEGAN, and Graph2Edits on the USPTO-50K dataset. The results showed that State2Edits outperformed the other semi-template-based methods in terms of top 1 and top 3 exact match accuracy, demonstrating the predictive efficiency of the model. Despite the promising performance, its limitations include the tendency of smaller motifs to produce longer edit sequences and subsequently reduce prediction accuracy, and the possibility that the aggregation method may result in incorrect predictions biased by functional groups. Ye et al. conclude that future work will focus on improving the generation of chemically coherent structures, including reactivity assessment, and making more effective use of reaction class information.  

\section{New Drug Design}\label{sec:newdrug}

The final category in this survey is new drug design, namely generative models that include GNNs. Although generative models have become quite popular for a variety of applications including drug discovery and molecule generation, the integration of GNNs in their frameworks seems to be less explored. In this section, three relevant papers are reviewed for their generative models involving GNNs applied to drug discovery. 

In 2021, a sequential molecular graph generator based on GNN, named MG$^2$N$^2$, was introduced by Bongini et al. \cite{MG2N2}. MG$^2$N$^2$ combines three GNN modules to solve node generation, edge classification, and edge generation in sequence for the task of unconditional molecular graph generation. The generation batches were evaluated in terms of validity, uniqueness, novelty, and VUN along with other generative models such as ChemVAE, MPGVAE, GrammarVAE, GraphVAE, MolGAN, and CGVAE on the QM9 and Zinc datasets. The proposed model was also evaluated in terms of QED, logP, and molecular weight over various configurations. While many of the previous models include Variational Autoencoders (VAE) that learn probabilistic representations of SMILES strings to generate similar SMILES strings, the sequential methodology of MG$^2$N$^2$ allows for greater interpretability as it is easy to identify where errors occur in the sequence. The results showed that MG$^2$N$^2$ outperformed many of the models in terms of VUN and consistently generated novel molecules, demonstrating impressive performance. Bongini et al. conclude that future work will focus on generalizing the model to other problems involving molecular graph generation and extend the approach to conditional molecule generation where ideal properties are included as inputs or constraints.  

Another generative model was proposed by Manu et al. in 2024, called GraphGANFed \cite{GraphGANFed}. The model implements GCN alongside a Generative Adversarial Network (GAN), which uses a generator to create new data samples while a discriminator tries to distinguish the generated samples from the real dataset samples, and Federated Learning (FL), which allows the model to be trained on different clients with unshared local datasets to preserve data privacy. Datasets ESOL, QM8, and QM9 were used to evaluate the model in terms of validity, uniqueness, novelty, internal diversity, QED, logP, and SNN. The results showed that the model consistently generated molecules with high novelty and diversity, and some additional takeaways were that a lower complexity discriminator model can help avoid mode collapse with smaller datasets, different metric values have a tradeoff that may depend on the dataset size or sample distribution, and the dropout ratio can be adjusted to modify the complexity of the generator and discriminator models. 

The MedGAN model proposed by Macedo et al. was another generative model from 2024, similarly combining a Wasserstein Generative Adversarial Network (WGAN) with a GCN \cite{MedGAN}. Unlike the other more generally applied models, MedGAN was created specifically to generate quinoline-scaffold molecules, which are promising drug candidates for their distinct chemical properties and potential for anticancer, anti-inflammatory, antibacterial, and antiviral activities. The model was optimized using a subset of PubChem and fine-tuned using ZINC15, with a focus on finding the most optimal model configuration rather than a typical comparison with baseline models. In terms of validity, connected validity, quinoline scaffold, novelty, diversity, number of unique and novel quinolines, Lipinski rule of 5, toxicity, and synthetic accessibility, the best performing configuration had a latent space of 256, Tanh/ReLU activation, and RMSProp optimization. With this, MedGAN was able to generate thousands of valid, novel, and unique quinoline molecules with desirable properties, while also demonstrating the robustness of the model even without components such as reinforcement learning or attention mechanisms. Still, some major limitations of GAN-based models include lack of interpretability and high computational requirements, which will need to be addressed in future work. 

\section{Future Work}\label{sec:futurework}
Although there has been great progress toward making effective use of GNN in drug discovery, there are still several frontiers for future work, many of which apply to multiple categories. One limitation that seems to appear quite frequently is simply the lack of high-quality, diverse, or specific data, which suggests the need for either more data to be collected/synthesized or models that make better use of the data that is available. There have been many efforts in the research to address this, such as considering somewhat related data that may not be as direct or even generating and using synthetic data. Still, there are improvements to be made in terms of data and its effective use in each category, which makes it a valid idea for future work.  

Another avenue for future work would be to continue optimizing GNN architectures, including the backbone algorithms themselves and the proposed models. Some ideas involve combining methods and models that provided exceptional performance in previous work to leverage the benefits of effective components, though this could also be explored by experimenting further with hybrid methods that make use of other algorithms, especially in the interest of creating a more accurate model with less computational complexity. Reviewing the literature may provide specific inspiration for these types of innovations, and experimentation is needed to verify the efficacy of proposed models.  

There has been significant work toward interpretable models for GNNs in drug discovery, as interpretability typically increases trust and allows experts to analyze the outputs of predictive models, though many models are still weak in this. Some future directions would be to include visualizations of predictions, especially since this type of data is graphical, and provide detailed attention-driven explanations. Although some models have integrated these concepts already, there should potentially be more emphasis on effective interpretability for any proposed model.  

In terms of less explored directions, a few categories were notably sparser than others. As mentioned, the use of GNNs has been seemingly overlooked in generative methods for new drug design. Since GNNs have been shown to improve model performance across categories, it would be reasonable to further experiment with their integration in generative models. Additionally, microbiome interaction is still an intriguing direction for future work, as it can be approached in several different ways. While there has been a decent amount of work in using GNNs for the prediction of microbe-drug associations, there is still not as much from the perspective of drug susceptibility to depletion in the human microbiome, besides the contribution from Maryam et al. Beyond that, there are surely additional drug interactions within the complex environment of the human body to potentially be explored with GNNs, which may be difficult to approach due to lack of applicable data.  

There are many other specific ideas for future work to be found within the reviewed work, though these are just a few general ideas for improvement that have impacted all categories. With a concept that has been explored increasingly in recent years, there is certainly no lack of suggestions and directions for improvement. With innovative future work, GNNs have the potential to become even more effective in real-world application. 

\section{Conclusions}\label{sec:conclusion}
In this paper, a total of 38 impactful and recent papers proposing GNN models for drug discovery across seven categories were reviewed, along with four additional review papers. The reviewed literature highlights the potential and proof of GNNs as an effective computational aid for the complex process of drug discovery, covering threads such as molecular property prediction, drug-target binding affinity, drug-drug interaction and synergy, microbiome interaction, drug repositioning, retrosynthesis, and new drug design. Although there has been great progress made toward improving GNN models for this task, even the most recent work reveals that there is still room for improvement. The current limitations of state-of-the-art models provide ideas and inspiration for continued optimization and exploration, and future work has a promising outlook with these directions. 

\bibliographystyle{IEEEtran}
\bibliography{refs}


%

\ifCLASSOPTIONcaptionsoff
  \newpage
\fi

\end{document}